\documentclass[sigconf]{acmart}
\AtBeginDocument{%
  \providecommand\BibTeX{{%
    \normalfont B\kern-0.5em{\scshape i\kern-0.25em b}\kern-0.8em\TeX}}}

\copyrightyear{2021}
\acmYear{2021}
\setcopyright{acmcopyright}\acmConference[ICS '21]{2021 International Conference on Supercomputing}{June 14--17, 2021}{Virtual Event, USA}
\acmBooktitle{2021 International Conference on Supercomputing (ICS '21), June 14--17, 2021, Virtual Event, USA}
\acmPrice{15.00}
\acmDOI{10.1145/3447818.3460372}
\acmISBN{978-1-4503-8335-6/21/06}

\usepackage{amsmath}
\usepackage{algorithm}
\usepackage[noend]{algorithmic}
\usepackage{multirow}
\usepackage{pgfplots}
\usepackage{tikz}
\usepgfplotslibrary{groupplots}
\pgfplotsset{compat=1.15}

\setcopyright{none}

\begin{document}

\title{Partitioning Sparse Deep Neural Networks for Scalable Training and Inference}

\author{Gunduz~Vehbi~Demirci}
\email{gunduz.vehbi.demirci@gmail.com}
\affiliation{%
\institution{University of Warwick}
\country{United Kingdom}
}

\author{Hakan~Ferhatosmanoglu}
\email{hakan.f@warwick.ac.uk}
\affiliation{%
\institution{University of Warwick}
\country{United Kingdom}
}

\renewcommand{\shortauthors}{Demirci and Ferhatosmanoglu}

\begin{abstract}
The state-of-the-art deep neural networks~(DNNs) have significant computational and data management requirements. The size of both training data and models continue to increase. Sparsification and pruning methods are shown to be effective in removing a large fraction of connections in DNNs. The resulting sparse networks present unique challenges to further improve the computational efficiency of training and inference in deep learning.  Both the feedforward~(inference) and backpropagation steps in stochastic gradient descent~(SGD) algorithm for training sparse DNNs involve consecutive sparse matrix-vector multiplications~(SpMVs).
We first introduce a distributed-memory parallel SpMV-based solution for the SGD algorithm to improve its scalability.
The parallelization approach is based on row-wise partitioning of weight matrices that represent neuron connections between consecutive layers.
We then propose a novel hypergraph model for partitioning weight matrices to reduce the total communication volume and ensure computational load-balance among processors. 
Experiments performed on sparse DNNs demonstrate that the proposed solution is highly efficient and scalable. By utilizing the proposed matrix partitioning scheme, the performance of our solution is further improved significantly.
\end{abstract}



\maketitle

\section{Introduction}

Deep neural networks~(DNNs) have been extensively utilized in computer vision, speech recognition, and natural language processing~\cite{krizhevsky2012imagenet, graves2005framewise, collobert2011natural}.
The state-of-the-art DNN architectures demand high storage and computational resources due to the large numbers of parameters~(i.e., connection weights) trained over huge datasets. 
For instance, AlexNet~\cite{krizhevsky2012imagenet}, Deepface~\cite{taigman2014deepface}, VGG16~\cite{simonyan2014very} and GPT-3~\cite{brown2020language} consist of $60$M, $120$M, $138$M and $175$B parameters, respectively. As both the number of parameters and the size of training datasets continue to increase, it is essential to develop scalable training and inference solutions.

Neural network pruning and sparsification methods are successfully applied to address the storage and computational challenges of DNNs~\cite{kepner2020graphchallenge,lecun1990optimal, hassibi1993second,srivastava2014dropout,liu2015sparse,geng2019o3bnn}.
These approaches aim at reducing the amount of memory and computation required to propagate values through the network, typically by removing unimportant connections. They improve DNN's efficiency, scalability, and feasibility in practice, especially for dynamic applications with low latency requirements~\cite{zhu2017prune}. Research studies demonstrate that DNNs are tolerant to the sparsification process~\cite{hooker2019selective,louizos2017learning}.
For instance, removal of $90\%$ of the connections in ResNet-50~\cite{he2016deep} incurs only $3\%$ accuracy loss~\cite{gale2019state}, when trained over ImageNet~\cite{deng2009imagenet}.

Stochastic gradient descent~(SGD) is a widely used method for training DNNs.
To achieve large-scale learning tasks, parallel SGD algorithms for distributed computing systems~(e.g., HPC systems, GPU clusters, TPU pods) are considered in the  literature~\cite{castello2019analysis, li2014scaling,awan2017s,coates2013deep,dean2012large,zhang2015deep,you2019fast}.
SGD algorithms that exploit sparsity patterns of networks should be developed to attain efficient training of sparse DNNs and retraining of pruned DNNs.
Inference~(feedforward) and backpropagation phases of SGD involve consecutive matrix-vector multiplications in such a way that the output vector of one layer is fed as input to the next layer.
Matrices in each layer store connection weight parameters between neurons and are updated during the course of training.
In the case of sparse DNNs, these matrices become sparse so that computations in each layer heavily depend on sparse matrix-vector multiplications~(SpMV).

For large-scale sparse DNNs, we introduce a distributed-memory parallel SGD solution based on efficient parallelization of SpMVs performed in feedforward and backpropagation phases.
To perform parallel SpMVs in each layer, matrices and input-output vectors are row-wise partitioned among processors.
This partitioning strategy achieves model-wise parallelism. This is in contrast to data-parallel approaches which necessitate the entire model to be stored by processors and face high bandwidth costs and memory bottleneck to perform parameter updates~\cite{wangni2018gradient}.
Our solution reduces memory requirements and performs efficient parameter updates via model-wise parallelization and utilizes sparse point-to-point communication operations to alleviate bandwidth and latency costs.

We then propose a hypergraph model for partitioning matrices to further scale and improve the efficiency of parallel SpMV computations by reducing the communication costs and achieving computational load-balance among processors. 
The proposed model utilizes partitioning with fixed vertices to correctly encode the communication requirements of processors and dependencies between successive layers.
The partitioning objective of minimizing the cut size in the hypergraph directly encodes the minimization of the total communication volume, and load-balancing constraints enable computational balance among processors.

To evaluate the performance of the proposed training solution with the hypergraph partitioning model, we conduct extensive experiments on several sparse DNN models provided by the Sparse Deep Neural Network Graph Challenge~\cite{kepner2019sparse} and the MNIST database of handwritten digits~\cite{lecun1998mnist}.
Experimental results show that the parallel SpMV-based sparse DNN training algorithm is highly efficient and scalable, and scales to large processor counts, and the proposed hypergraph partitioning model provides further performance improvements and scalability by significantly reducing both the bandwidth and latency costs of communication.

The contributions of the paper are as follows:
\begin{itemize}
\item We introduce a distributed memory-parallel SGD algorithm specifically designed for sparse DNNs to achieve model-wise parallelism.
\item To improve parallelization efficiency, we propose a novel hypergraph-based sparse DNN partitioning model which reduces communication costs and achieves a computational balance among processors.
\item On a set of sparse DNNs from a benchmark comprising realistic representatives of real-world applications, we performed extensive experiments to analyze the scalability and effectiveness of the proposed algorithm and partitioning model.

\end{itemize}

The rest of the paper is organized as follows.
Section~\ref{sec-related} presents related work.
Section~\ref{sec:prelims} presents preliminaries.
Section~\ref{sec:sparseSGD} describes the proposed distributed-memory parallel SpMV-based SGD solution for sparse DNNs.
Section~\ref{sec-HSpFF} describes our hypergraph model for partitioning sparse DNNs.
Section~\ref{sec-Performance} presents experimental results for performance evaluation.
Finally, Section~\ref{sec-conclusion} concludes the paper.

\section{Related Work}\label{sec-related}

Efficient parallel SpMV algorithms for distributed-memory and shared-memory systems are developed in the literature~\cite{schubert2011parallel,akbudak2013hypergraph,yang2011fast}.
Several graph/hypergraph partitioning models are proposed to improve the performance of parallel SpMV by reducing the communication costs and achieving the load-balance among processors~\cite{kaya2015scalable, catalyurek1999hypergraph,hendricksonpartitioning,kolda1998partitioning}.
Existing approaches, however, are suitable mostly for the cases in which an input matrix is repeatedly multiplied by a vector where the sparsity pattern of the input matrix does not change through the iterations.
Hence, these partitioning models and parallel SpMV algorithms are not applicable for sparse DNNs, since each layer is associated with a sparse matrix with different nonzero patterns.
Research is needed to design solutions that address the challenges introduced by sparse DNNs and improve their performance.

Motivated by the computational advantages and reduced sizes to handle very large data and models, efficient inference computation on sparse DNNs has attracted significant attention~\cite{kepner2019sparse}.
Parallel algorithms for sparse computations on shared-memory systems are recently proposed~(e.g., GPUs~\cite{bisson2019gpu,wang2019accelerating,ning2019deep,hidayetouglu2020scale}, multiprocessors~\cite{davis2019write,mofrad2019multithreaded, pawlowski2020combinatorial}).
Since these approaches implement only inference computation and are not used for training, each input data vector can be independently processed and distributed parallelism can be achieved by just splitting the input dataset and replicating DNN models among multiple compute nodes.
Recently, novel tiling strategies for sparse DNNs are developed to utilize dense matrix kernels for GPUs ~\cite{guo2020accelerating}.

Data-parallel methods are widely used to achieve scalability via distributed SGD.
In these methods, the dataset is partitioned among multiple compute nodes and local portions of the dataset are processed in terms of batches.
Additionally, each compute node stores a local copy of the whole DNN model and depending on the implementation, synchronous or asynchronous updates are performed on the model parameters.
Data-parallel SGD algorithms necessitate a large volume of communication between processors since whole model parameters are transferred at each iteration. 
Therefore, to make data-parallel approaches more feasible, the batch size needs to be increased, but larger batch sizes hurt the training performance of the SGD algorithm.
Additionally, to alleviate the high communication cost, gradient compression methods are proposed~\cite{aji2017sparse,lin2017deep}.
More recently, FFT-based gradient sparsification and range-based quantization methods are applied together to reduce the communication volume in data-parallel training algorithms~\citep{wang2020fft}.

In synchronous data-parallel methods~\cite{li13pytorch,iandola2016firecaffe,goyal2017accurate,iandola2016firecaffe,you2019fast,awan2017s}, each node computes local gradients independently and all processors collectively perform an All-reduce communication to receive the average of gradients to update its local parameters.
Recently, communication algorithms to improve the efficiency of All-reduce operation on NVLink-enabled dense GPU systems are proposed~\cite{chu2020nv}.
To achieve efficiency in synchronous SGD, larger batch sizes should be considered, which may result in lower test accuracy.
Methods are proposed to reduce the loss of accuracy due to the use of larger batch sizes~\cite{goyal2017accurate,you2017scaling}.

The requirement for processors to synchronize gradient updates after processing each batch causes a significant limitation for the scalability of synchronized SGD.
Techniques to overlap communication and computation are proposed to reduce the overheads of synchronization~\cite{goyal2017accurate,das2016distributed}.
To achieve further performance improvements, asynchronous methods which differ in communication and update rules are proposed~\cite{chilimbi2014project,dean2012large,zhang2015deep,jin2016scale}.
In asynchronous methods, at each step, a master node~(i.e., parameter server) receives local gradients from a worker node, and updates global model parameters then sends the updated model to the same worker where worker nodes are served in arbitrary order.
Federated learning algorithms~\cite{konevcny2016federated,chai2020tifl} are also considered under this category where a subset of clients download the most recent model from a central server and computes updates to the model.
Then the clients send their model updates to the central server which aggregates these model updates typically by averaging to improve the global model.

Alternative to data-parallel methods, training approaches that aim at model-wise parallelism are also considered~\cite{jia2018exploring, jia2019beyond}.
For example, FlexFlow~\cite{jia2019beyond} searches different parallelization strategies by performing simulations before training.
However, this tool is mainly designed for GPU clusters and does not provide a partitioning on sparse DNNs.
The proposed model-wise parallelism in our SGD algorithm offers inherent scalability, whereas in data-parallel approaches, each processor holds the whole set of parameters and broadcasts gradients for these parameters to all processors.
Therefore, in data-parallel approaches, the total communication volume significantly increases with increasing number of processors, and the local memory size of processors limits the size of neural networks.
As validated in the experiments, the proposed SGD algorithm reduces the total communication volume, since each processor only keeps a small set of parameters and broadcasts their gradients to a small subset of processors.

\section{Preliminaries}\label{sec:prelims}
\subsection{Hypergraph Partitioning}\label{sec-hgp}

Let $H\!=\!(V,N)$ denote a hypergraph where $V$ and $N$ are vertex and net sets, respectively.
Each net $n_j\!\in\!N$ may connect multiple vertices and the set of vertices that connected by $n_j$ is represented by $\mathrm{pins}(n_j)$.
Each vertex $v_i\!\in\!V$ is associated with weight $w(v_i)$ and each net $n_j\!\in\!N$ is associated with $\mathrm{cost}(n_j)$.
A $P$-way partition of $H$ is defined as $\Pi\!=\!\{V_1, V_2 \cdots V_P\}$ consisting of mutually disjoint and exhaustive subsets of vertices $V_m\!\subset\!V$ where $V_m\cap V_n\!=\!\emptyset$ if $m\!\neq\!n$ and $V_m\!\neq\!\emptyset$ for all $V_m\!\in\!\Pi$ such that $\bigcup V_m = V$.

Under a partition $\Pi$, a net $n_j$ connects to a part $V_m$ if $\mathrm{pins}(n_j)\!\cap\!V_m\!\neq\!\emptyset$.
The set of parts that are connected by net $n_j$ is defined as the connectivity set $\Lambda(n_j)$ and the number of parts that are connected by net $n_j$ is defined as connectivity $\lambda(n_j)\!=\!|\Lambda(n_j)|$.
A net $n_j$ is said to be cut if it connects to multiple parts~(i.e., $\lambda(n_j)\!>\!1$) and uncut otherwise.
The connectivity cut size under $\Pi$ is defined as
\begin{equation}
\chi(\Pi) = \sum\limits_{n_j \in N} \mathrm{cost}(n_j)\times(\lambda(n_j) -1)
\label{equation:cutsize}
\end{equation}
The weight of a part $V_m\!\in\!\Pi$ is defined as $W(V_m)\!=\!\sum_{v_i \in V_m}w(v_i)$.
The partition $\Pi$ is balanced if it satisfies
\begin{equation}
W(V_m) \leq W_{avg} (1 + \epsilon), \,\,\,\, \forall \, V_m \in \Pi 
\label{equation:balancing}
\end{equation}
where ${W_{avg} = \sum_{v_i \in V}w(v_i) / P}$ is the average part weight and $\epsilon$ is the maximum allowed imbalance ratio.

The hypergraph partitioning problem for finding a $P$-way partition with the objective of minimizing the cut size given in~(\ref{equation:cutsize}) and satisfying balancing constraints in~(\ref{equation:balancing}) is NP-Hard.
There exist tools that produce quality results for the hypergraph partitioning problem~\cite{catalyurek1999hypergraph,karypis1998hmetis}.
These tools also support partitioning hypergraphs with fixed vertices where some vertices can be assigned to parts prior to partitioning.

\subsection{Stochastic Gradient Descent}
\label{sec-GD}
Stochastic gradient descent~(SGD) is an optimization technique which is commonly used for training neural networks to iteratively minimize a loss function over an input dataset.
SGD is usually implemented in two main phases which heavily depend on SpMVs: (1)~Feedforward~(inference) phase, (2)~Backpropagation phase.

Given a DNN composed of $L$ layers where connection weights in each layer $k\!=\!1,2,\ldots,L$ are represented by a matrix $\textbf{W}^{k}$ such that the connection weight from the $i$th neuron in layer $k$ to the $j$th neuron in layer $k\!+\!1$ is denoted by nonzero entry $\textbf{W}^{k}(j,i)$.
In the inference phase, an input vector $\textbf{x}^{0}$ is sent through the network layers to compute an output vector $\textbf{x}^{L}$. 
Formally, the inference step can be given as
\begin{eqnarray}
\textbf{x}^{k} = f(\textbf{W}^{k} \textbf{x}^{k-1} + \textbf{b}^{k})
\label{eq:feedforward}
\end{eqnarray}
\noindent 
where $\textbf{b}^{k}$ denotes the bias vector and $f(\cdot)$ is a nonlinear activation function applied to each element of a vector. 
The bias vector $\textbf{b}^{k}$ can be embedded in matrix $\textbf{W}^{k}$ as the first column and the first entries of vectors $\textbf{x}^{k}$ can be set to one~(i.e., the number of dimension of $\textbf{x}^{k}$ increases by one). 
In a simpler form, the feedforward computation in each layer $k$ becomes $\textbf{x}^{k}\!=\!f(\textbf{W}^{k} \textbf{x}^{k-1})$.

In the backpropagation phase, output vector $\textbf{x}^{L}$ of the inference step is used for computing gradient vector $\boldsymbol{\delta}^{L}$ which is backpropagated to compute gradients $\boldsymbol{\delta}^{k}$ in preceding layers $k=1,2,\ldots,L\!-\!1$.
The $i$th component of vector $\boldsymbol{\delta}^{k}(i)$ denotes the partial derivative of a loss function $\textbf{J}(\textbf{x}^{L}, \textbf{y})$ with respect to the total input activation of the $i$th neuron in layer $k$.
Vector $\textbf{y}$ is the true label for input vector $\textbf{x}^0$ where the loss function depends on both of the vectors.
Each gradient $\boldsymbol{\delta}^{k}$ is used to update weight matrix $\textbf{W}^{k}$ by the following gradient update rule
\begin{eqnarray}
\frac{\partial \textbf{J}}{\partial \textbf{W}^{k}(j,i)} = \boldsymbol{\delta}^k(j) \, \textbf{x}^{k-1}(i) \\
\textbf{W}^{k}(j,i) \leftarrow \textbf{W}^{k}(j,i) - \eta \frac{\partial \textbf{J}}{\partial \textbf{W}^{k}(j,i)}
\end{eqnarray}
\noindent where $\eta$ denotes the learning rate.
The gradient vector $\boldsymbol{\delta}^{L}$ in the final layer $L$ is computed as
\begin{equation}
\boldsymbol{\delta}^{L} = \nabla_{\textbf{x}^{L}} \textbf{J} \ \odot f^{\prime}(\textbf{z}^{L})\label{equ:final-gradient}
\end{equation}
\noindent
where $\nabla_{\textbf{x}^{L}} \textbf{J}$ is a vector of derivatives of the loss function $\textbf{J}$ with respect to the outputs of the activation functions in the final layer~(i.e., $\textbf{x}^{L}$) and $f^{\prime}(\textbf{z}^{L})$ is the vector of derivatives of the outputs with respect to the input activation $\textbf{z}^{L}= \textbf{W}^{L} \textbf{x}^{L}$~(i.e., local gradients) in layer $L$ and symbol ``$\odot$'' denotes element-wise multiplication.  
Gradients for layers $k=1,2,\ldots,L\!-\!1$ are computed by a recursive formula 
\begin{equation}
\boldsymbol{\delta}^{k-1} = (\textbf{W}^{k})^T \boldsymbol{\delta}^{k} \ \odot f^{\prime}(\textbf{z}^{k-1}).
\end{equation}

\begin{algorithm}
\begin{algorithmic}[1]
\REQUIRE $T$, $\{\mathbf{W}^{k}\}$
\FOR{$\textbf{each}$ $\textbf{x}^{0} \in T$}    
    \FOR{$k=1,2,\ldots, L$} 
    \STATE {$\textbf{z}^{k}\!=\!\textbf{W}^{k} \textbf{x}^{k-1}$} 
    \STATE {$\textbf{x}^{k}\!=\!f(\textbf{z}^{k})$} 
    \ENDFOR
    \STATE {$\boldsymbol{\delta}^{L} = \nabla_{\textbf{x}^L} J \ \odot f^{\prime}(\textbf{z}^L)$} 
    \FOR{$k=L,L\!-\!1,\ldots, 1$} 
    \STATE {$\boldsymbol{\delta}^{k-1} = (\textbf{W}^{k})^T \boldsymbol{\delta}^{k} \ \odot f^{\prime}(\textbf{z}^{k-1})$} 
    \STATE {$\nabla\textbf{W}^{k} =  \boldsymbol{\delta}^k \otimes \textbf{x}^{k-1}$} 
    \STATE {$\textbf{W}^{k} \leftarrow \textbf{W}^{k} - \eta \nabla\textbf{W}^{k}$} 
    \ENDFOR
\ENDFOR    
 
\end{algorithmic}
\caption{SGD}\label{alg:sgd}
\end{algorithm}

Algorithm~\ref{alg:sgd} displays the overall execution of SGD.
The for loop in lines~$1$--$9$ is executed overall input vectors in training dataset $T$ in such a way that for each input vector $\textbf{x}^{0}\!\in\!T$, feedforward and backpropagation steps are executed.
Lines~$2$--$4$ correspond to inference~(feedforward) step where repeated SpMVs of the form $\textbf{z}^{k}\!=\!\textbf{W}^{k} \textbf{x}^{k-1}$ are performed.
Between two consecutive layers, nonlinear activation function $f(\cdot)$ is applied to each component of vector $\textbf{z}^{k}$ and the output vector $f(\textbf{z}^{k})$ of layer $k$ is fed as input to the next layer $k\!+\!1$.
In line~$5$, the gradient vector $\boldsymbol{\delta}^L$ is computed using the output vector $\textbf{x}^L$ and the input activations $\textbf{z}^L$ of the final layer $L$.
Lines~$6$--$9$ correspond to backpropagation step where  repeated SpMVs of the form $\boldsymbol{\delta}^{k-1}\!=\!(\textbf{W}^{k})^T \boldsymbol{\delta}^{k}$ are performed to backpropagate gradient vectors.
In line~$8$, outer product of gradient vector $\boldsymbol{\delta}^{k}$ with vector $\textbf{x}^{k-1}$ is performed to produce matrix $\nabla\textbf{W}^{k}$ which is used to update weight matrix $\textbf{W}^{k}$ by the gradient update rule.

\section{Distributed SGD algorithm for Sparse DNNs}\label{sec:sparseSGD}

In order to achieve a parallel training algorithm (i.e., parallel SGD) for sparse DNNs, we develop parallel SpMV-based feedforward and backpropagation algorithms in Sections~\ref{sec-SpFF} and~\ref{sec-SpBP}, respectively.
That is, parallel sparse feedforward~(SpFF) algorithm achieves parallelization of lines~2--4, whereas parallel sparse backpropagation~(SpBP) algorithm achieves parallelization of lines~5--9 in Algorithm~\ref{alg:sgd}.

\begin{figure}[t]
\centering
\includegraphics[width=0.49\textwidth]{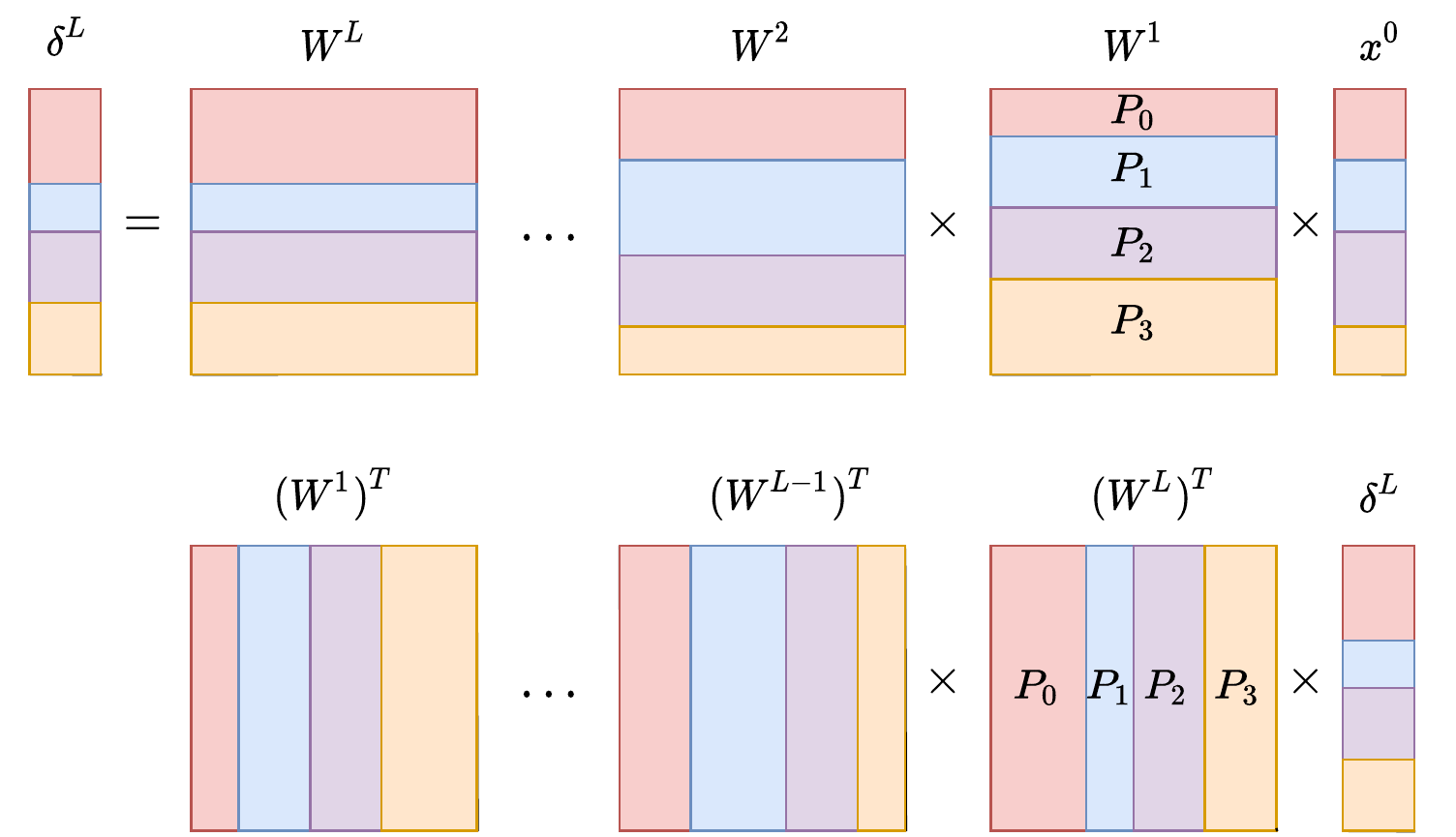}
\caption{Illustration of sequences of SpMVs performed in SpFF~(Top) and SpBP~(Bottom).}
\label{fig-sgd-spmv}
\end{figure}

Figure~\ref{fig-sgd-spmv} displays the general execution of the parallel SGD algorithm together with its weight matrix partitioning scheme.
In the figure, only sequences of SpMV operations are displayed whereas the remaining computations are omitted for ease of exposition.
In the inference phase, input vector $\textbf{x}^0$ and weight matrices $\textbf{W}^1, \textbf{W}^2 \ldots \textbf{W}^L$ are row-wise partitioned among four processors.
For each layer $k\!=\!1,2,\ldots L$, processors perform communication to receive non-local entries of vector $\textbf{x}^{k-1}$ and collectively perform SpMV $\textbf{W}^k \textbf{x}^{k-1}$ to compute vector $\textbf{x}^{k}$.
The output vector $\textbf{x}^{k}$ computed in layer $k$ is used as input in the next layer.
This process is repeatedly performed until the final layer $L$ where the gradient vector $\boldsymbol{\delta}^{L}$ is computed.
In the backpropagation phase, gradient vector $\boldsymbol{\delta}^{L}$ is row-wise partitioned among processor whereas transposes of weight matrices $(\textbf{W}^1)^T, (\textbf{W}^2)^T \ldots (\textbf{W}^L)^T$ are column-wise partitioned.
The row-wise partitioning of weight matrices induces column-wise partitioning on their transposes.
For each layer $k\!=\!L,L\!-\!1,\ldots 1$, processors collectively perform SpMV $(\textbf{W}^{k})^T \boldsymbol{\delta}^{k}$.
Here, since matrices are column-wise partitioned, processors communicate partial products contributing to the same nonzero entries of output vector $\boldsymbol{\delta}^{k-1}$ instead of communicating entries of input vector $\boldsymbol{\delta}^{k}$.
These partial products are summed by processors to get the final values of entries in gradient vector $\boldsymbol{\delta}^{k-1}$ which is used as input in the next layer.

\subsection{Parallel Sparse Feedforward}\label{sec-SpFF}

The parallel sparse feedforward~(SpFF) performs repeated parallel SpMV in the form of $\textbf{W}^{k}\textbf{x}^{k-1}$ for each layer $k\!=\!1,2,\ldots,L$. 
Parallelism is achieved through row-wise partitioning of weight matrices $\mathbf{W}^{k}$ and input/output vectors $\mathbf{x}^{k-1}$ among processors.

Algorithm~\ref{alg:spff} displays the overall execution of the proposed SpFF algorithm.
In the algorithm, each processor $P_m$ for $m\!=\!1,2,\ldots,P$ stores row-blocks $\mathbf{W}^{k}_{m}$ and $\mathbf{x}^{k-1}_{m}$ of matrix $\mathbf{W}^{k}$ and vector $\mathbf{x}^{k-1}$, respectively.
Additionally, each processor $P_m$ is provided with maps $\mathrm{Xsend}^{k}_m$ and $\mathrm{Xrecv}^{k}_m$ that map row indices of vector $\mathbf{x}^{k-1}_{m}$ to processor ids.
In this way, each processor knows which $\mathbf{x}^{k-1}$-vector entries to be communicated with which processor.
Formally, these sets are defined as
\begin{eqnarray}
\begin{split}
&\mathrm{Xsend}^{k}_m = \left\lbrace (P_n,\mathbf{\bar{x}}^{k-1}_{mn}) \mid \right. \\
&\left. \mathbf{\bar{x}}^{k-1}_{mn} = \mathbf{x}^{k-1}_{m} [ cols(\mathbf{W}^{k}_{n}) \cap rows(\mathbf{W}^{k-1}_{m})]  \right\rbrace  
\end{split}\label{eq2:1}\\
\begin{split}
&\mathrm{Xrecv}^{k}_m = \left\lbrace (P_n,\mathbf{\hat{x}}^{k-1}_{nm}) \mid \right. \\
&\left. \mathbf{\hat{x}}^{k-1}_{nm} = \mathbf{x}^{k-1}_{n}[ cols(\mathbf{W}^{k}_{m}) \cap rows(\mathbf{W}^{k-1}_{n}) ] \right\rbrace\label{eq2:2}
\end{split}
\end{eqnarray}
\noindent
where $cols(\cdot)$ and $rows(\cdot)$ respectively denote the indices of columns and rows that contain at least one nonzero entry in a given matrix/vector. $\mathbf{x}^{k-1}_{m}[\cdot]$ and $\mathbf{x}^{k-1}_{n}[\cdot]$ denote subvectors that are composed of given row indices of vectors $\mathbf{x}^{k-1}_{m}$ and $\mathbf{x}^{k-1}_{n}$, respectively.
Hence, for each $(P_n,\mathbf{\bar{x}}^{k-1}_{mn})\!\in\! \mathrm{Xsend}^{k}_m$, processor $P_m$ sends subvector $\mathbf{\bar{x}}^{k-1}_{mn}$ to processor $P_n$ whereas for each $(P_n,\mathbf{\hat{x}}^{k-1}_{nm})\!\in\!\mathrm{Xrecv}^{k}_m$, processor $P_m$ receives subvector $\mathbf{\hat{x}}^{k-1}_{nm}$ from processor $P_n$.

\begin{algorithm}[t]
\begin{algorithmic}[1]
\REQUIRE $\mathbf{x}^{0}_m$, $\{\mathbf{W}^{k}_m\}$, $\{\mathrm{Xsend}^{k}_m\}$, $\{\mathrm{Xrecv}^{k}_m\}$
\FORALL{\emph{processors} $P_m$ \textbf{in parallel}}
 \FOR{$k=1,2,\ldots, L$} 
  \FOR{$\textbf{each}$ $(P_n,\mathbf{\bar{x}}^{k-1}_{mn}) \in \mathrm{Xsend}^{k}_m$} 
  \STATE{Update $\mathbf{\bar{x}}^{k-1}_{mn}$ with entries in $\mathbf{x}^{k-1}_{m}$}
  \STATE{Non-blocking send $\mathbf{\bar{x}}^{k-1}_{mn}$ to $P_n$}
  \ENDFOR
  \STATE{$\mathbf{z}^{k}_m = \mathbf{W}^{k}_{m}\mathbf{x}^{k-1}_{m}$}  
  
   \FOR{$\textbf{each}$ $(P_n,\mathbf{\hat{x}}^{k-1}_{nm}) \in \mathrm{Xrecv}^{k}_m$} 
  \STATE{Receive nonzero entries $\mathbf{\hat{x}}^{k-1}_{nm}$ from process $P_n$}
  \STATE{$\mathbf{z}^{k}_m \leftarrow \mathbf{z}^{k}_m + \mathbf{W}^{k}_{m}\mathbf{\hat{x}}^{k-1}_{nm}$}
  \ENDFOR
  \STATE{$\mathbf{x}^{k}_{m} = f(\mathbf{z}^{k}_m)$}
 \ENDFOR
\ENDFOR
\end{algorithmic}
\caption{SpFF}\label{alg:spff}
\end{algorithm}

Sets $\mathrm{Xsend}^{k}$ and $\mathrm{Xrecv}^{k}$ are precomputed by using the sparsity patterns of weight matrices~(i.e., neuron connections) and the row partitioning of weight matrices among processors. 
The row-wise partitioning of weight matrices induces neuron partitioning in each layer so that all computations related to a neuron are performed by a single processor.
As shown in~(\ref{eq2:1}) and~(\ref{eq2:2}), to perform $\mathbf{W}^{k}_{m}\mathbf{x}^{k-1}$, processor $P_m$ needs to receive all $\mathbf{x}^{k-1}$-vector rows corresponding to column indices in $cols(\mathbf{W}^{k}_{m})$.
It is important to note that vectors $\mathbf{\bar{x}}^{k-1}_{mn}$ and $\mathbf{\hat{x}}^{k-1}_{nm}$ are placeholders that keep coordinates of nonzero entries.
Hence, nonzero entries of these vectors are updated before used in any operation.
For instance, before sending to processor $P_n$, nonzero entries of vector $\mathbf{\bar{x}}^{k-1}_{mn}$ are updated~(i.e., line~4) with the corresponding entries in $\mathbf{x}^{k-1}_{m}$ locally computed in the preceding layer.
Similarly, nonzero entries of vector $\mathbf{\hat{x}}^{k-1}_{nm}$ must be received from processor $P_n$ before it is multiplied by weight matrix $\mathbf{W}^{k}_{m}$~(i.e., lines~$8$--$9$).

In the algorithm, for each layer $k\!=\!1,2,\ldots,L$, the for loop in line~2 is executed in parallel by all processors: 
In lines~3--5, each processor $P_m$ performs a non-blocking communication for each tuple $(P_n,\mathbf{\bar{x}}^{k-1}_{mn})\!\in\!\mathrm{Xsend}^{k}_m$ to send its local nonzero entries in $\mathbf{\bar{x}}^{k-1}_{mn}$ to processor $P_n$.
To overlap communication by computation, each processor performs local SpMV computation $\mathbf{z}^{k}_m\!=\!\mathbf{W}^{k}_{m}\mathbf{x}^{k-1}_{m}$ without waiting for the messages to be received by recipient processors.
Entries of $\mathbf{z}^{k}$ store the total activation values incoming to neurons.
For instance, nonzero entry $\mathbf{z}^{k}(i)$ stores the total activation of the $i$th neuron in layer $k$.
After local SpMV computations are performed, for each tuple $(P_n,\mathbf{\hat{x}}^{k-1}_{nm})\!\in\!\mathrm{Xrecv}^{k}_m$, processor $P_m$ receives vector $\mathbf{\hat{x}}^{k-1}_{nm}$ from processor $P_n$ and multiplies by $\mathbf{W}^{k}_{m} $ to update the corresponding entries in vector $\mathbf{z}^{k}_m$~(i.e., lines~$7$--$9$).
Finally, a nonlinear activation function~(i.e., ReLu, sigmoid etc.) is applied to $\mathbf{z}^{k}_m$ and the respective output elements in $\mathbf{x}^{k}_{m}$ are computed.

\subsection{Parallel Sparse Backpropagation}\label{sec-SpBP}

The parallel sparse backpropagation~(SpBP) works similarly to SpFF algorithm where SpBP performs repeated SpMVs in the form of $\boldsymbol{\delta}^{k-1} = (\textbf{W}^{k})^T \boldsymbol{\delta}^{k}$ in the reverse order that of performed by SpFF.
Since the weight matrices are row-wise partitioned among processors, each processor $P_m$ for $m\!=\!1,2,\ldots,P$ stores column-block $(\mathbf{W}^{k}_{m})^T$ of matrix $(\mathbf{W}^{k})^T$ and row-block $\boldsymbol{\delta}^{k}_{m}$ of gradient vector $\boldsymbol{\delta}^{k}$, respectively.
Therefore, each processor $P_m$ multiplies its local gradient vector $\boldsymbol{\delta}^{k}_m$ by transpose $(\textbf{W}^{k}_m)^T$ of its local weight matrix $\textbf{W}^{k}_m$ in each layer~$k$.

Algorithm~\ref{alg:spbp} displays the overall execution of SpBP.
As a first step, each processor $P_m$ locally computes gradient vector $\boldsymbol{\delta}^L_m$ according to Eq.~(\ref{equ:final-gradient}) by using the output vector $\textbf{x}^L_m$ computed in the inference phase.
By executing the for loop in lines~$3$--$13$ in parallel, vector $\boldsymbol{\delta}^L$ is backpropagated through the layers $L,L\!-\!1,\ldots,1$.
To backpropagate vector $\boldsymbol{\delta}^{k}$ to the preceding layer $k\!-\!1$, an SpMV of the form $\textbf{s}^{k}_m\!=\!(\textbf{W}^{k}_m)^T \boldsymbol{\delta}^{k}_m$ is performed in line~$4$.
Vector $\textbf{s}^{k}_m$ may contain partial derivatives contributing to neuron outputs computed on different processors as well as to local neuron outputs.
Nonzeros of vector $\textbf{s}^{k}_m$ that are contributing to neurons located on different processors are sent to the corresponding processors.
Nonzeros that are contributing to the local neuron outputs are summed with the partial derivatives received from other processors, before multiplying with local gradients $f^{\prime}(\textbf{z}^{k-1})$.
That is, communication operations are performed on nonzero entries of $\textbf{s}^{k}_m$.

As in SpFF algorithm, each processor is provided with maps $\mathrm{Ssend}^{k}$ and $\mathrm{Srecv}^{k}$, where for each tuple $(P_n,\mathbf{\bar{s}}^{k}_{mn})\!\in\!\mathrm{Ssend}^{k}_m$ there exists $(P_n,\mathbf{\hat{x}}^{k-1}_{nm})\!\in\!\mathrm{Xrecv}^{k}_m$ and $rows(\mathbf{\bar{s}}^{k}_{mn})\!=\!rows(\mathbf{\hat{x}}^{k-1}_{nm})$.
Similarly, for each tuple $(P_n,\mathbf{\hat{s}}^{k}_{nm})\!\in\!\mathrm{Srecv}^{k}_m$ there exists $(P_n,\mathbf{\bar{x}}^{k-1}_{mn})\!\in\!\mathrm{Xsend}^{k}_m$ and $rows(\mathbf{\hat{s}}^{k}_{nm})\!=\!rows(\mathbf{\bar{x}}^{k-1}_{mn})$.
That is, if processor $P_m$ receives a nonzero $\mathbf{x}^{k-1}(i)$ from processor $P_n$, then $P_m$ sends the corresponding gradient contribution $\mathbf{s}^{k}(i)$ to $P_n$.
Similarly, if processor $P_m$ sends a nonzero $\mathbf{x}^{k-1}(j)$ to processor $P_n$, then $P_m$ receives the corresponding gradient contribution $\mathbf{s}^{k}(j)$ from $P_n$.

\begin{algorithm}[t]
\begin{algorithmic}[1]
\REQUIRE $\mathbf{x}^{L}_m$, $\{(\mathbf{W}^{k}_m)^T\}$, $\{\mathrm{Ssend}^{k}_m\}$, $\{\mathrm{Srecv}^{k}_m\}$
\FORALL{\emph{processors} $P_m$ \textbf{in parallel}}
\STATE{$\boldsymbol{\delta}^L_m = \nabla_{\mathbf{x}^L_m} \mathbf{J} \ \odot f^{\prime}(\mathbf{z}^L_m)
$}

 \FOR{$k=L,L\!-\!1,\ldots, 1$} 
 \STATE{$\mathbf{s}^{k}_m = (\textbf{W}^{k}_m)^T \boldsymbol{\delta}^{k}_m$}

  \FOR{$\textbf{each}$ $(P_n,\mathbf{\bar{s}}^{k}_{mn}) \in \mathrm{Ssend}^{k}_m$} 
  \STATE{Update $\mathbf{\bar{s}}^{k}_{mn}$ with corresponding entries in $\mathbf{s}^{k}_{m}$}
  \STATE{Non-blocking send $\mathbf{\bar{s}}^{k}_{mn}$ to $P_n$}
  \ENDFOR
  
  \STATE {$\nabla\textbf{W}^{k}_m = \boldsymbol{\delta}^{k}_m \otimes \textbf{x}^{k-1}_m$} 
  \STATE {$\textbf{W}^{k}_m \leftarrow \textbf{W}^{k}_m - \eta \nabla\textbf{W}^{k}_m$} 
 
  \FOR{$\textbf{each}$ $(P_n,\mathbf{\hat{s}}^{k}_{nm}) \in \mathrm{Srecv}^{k}_m$} 
  \STATE{Receive nonzero entries $\mathbf{\hat{s}}^{k}_{nm}$ from process $P_n$}
  \STATE{$\mathbf{s}^{k}_m \leftarrow \mathbf{s}^{k}_m + \mathbf{\hat{s}}^{k}_{nm}$}
  \ENDFOR
  
  \STATE{$\boldsymbol{\delta}^{k-1}_m = \mathbf{s}^{k}_m \ \odot f^{\prime}(\mathbf{z}^{k-1}_m)$}
  
 \ENDFOR
\ENDFOR
\end{algorithmic}
\caption{SpBP}\label{alg:spbp}
\end{algorithm}

In lines~$5$--$7$, each processor $P_m$ performs a non-blocking communication for each tuple $(P_n,\mathbf{\bar{s}}^{k}_{mn})\!\in\!\mathrm{Ssend}^{k}_m$ to send nonzero entries in $\mathbf{\bar{s}}^{k}_{mn}$ to processor $P_n$.
To overlap communication by computation, each processor locally performs outer product $\boldsymbol{\delta}^{k}_m\!\otimes \textbf{x}^{k-1}_m$ without waiting for the messages to be received by recipient processors.
It is important to note that $\textbf{x}^{k-1}_m$ contains nonzero entries received from other processors in the inference phase.
The outer product produces matrix $\nabla\textbf{W}^{k}_m$ which is used to update weight matrix $\textbf{W}^{k}_m$ in lines~$8$--$9$.
After updating weight matrices, for each tuple $(P_n,\mathbf{\hat{s}}^{k}_{nm})\!\in\!\mathrm{Srecv}^{k}_m$, processor $P_m$ receives nonzero entries in $\mathbf{\hat{s}}^{k}_{nm}$ from processor $P_n$ and 
sums the received nonzero entries with the corresponding entries in $\mathbf{s}^{k}_m$ to compute the final partial derivatives for the local neuron outputs~(i.e., lines~$10$--$12$).
Finally, nonzero entries of $\mathbf{s}^{k}_{m}$ are multiplied with local gradients in line~$13$ and gradient vector $\boldsymbol{\delta}^{k-1}_m$ for the preceding layer $k\!-\!1$ is computed.
It is important to highlight that only the nonzero entries of $\mathbf{s}^{k}_{m}$, which correspond to $rows(\mathbf{x}^{k-1}_{m})$, are multiplied by local gradients $f(\mathbf{z}^{k-1}_m)$ and carried into vector $\boldsymbol{\delta}^{k-1}_m$.

\section{Hypergraph Partitioning Model for sparse DNNs}\label{sec-HSpFF}

We propose a hypergraph model for partitioning rows of weight matrices~(i.e., neural network) among processors to optimize communication costs of parallel SpMV operations performed by SpFF and SpBP algorithms.
The proposed model adopts a multi-phase and fixed vertex partitioning approach to correctly encode communication patterns of processors between consecutive layers. 

Our partitioning model consists of $L$ phases $\phi^{k}$ for $k\!=\!1,2,\ldots,L$. 
In each phase $\phi^{k}$, rows of matrix $\mathbf{W}^{k}$ are partitioned into $P$ parts.
Note that the row-wise partitioning of weight matrix $\mathbf{W}^{k}$ induces column-wise partitioning of $(\mathbf{W}^{k})^T$ in backpropagation phase.
For each phase $\phi^{k}$, we define a hypergraph $H(\phi^{k})\!=\!(V^{k}\cup F^{k},N^{k})$, where for each matrix row $\mathbf{W}^{k}(i,:)$, there exists one vertex $v_i\!\in\!V^{k}$, for each column $\mathbf{W}^{k}(:,j)$, there exists one fixed vertex $v^f_j\!\in\!F^{k}$ and one net $n_j\!\in\!N^{k}$.

Each vertex $v_i\!\in\!V^{k}$ represents row $\mathbf{W}^{k}(i,:)$~(i.e., the $i$th neuron) and all computations associated with that row.
In the inference phase, vertex $v_i$ represents the task of computing the inner product 
\begin{equation}
\begin{split}
\mathbf{z}^{k}(i) = \mathbf{W}^{k}(i,:)\mathbf{x}^{k-1} =\hspace{-1em}\sum\limits_{\mathbf{W}^{k}(i,j) \in \mathbf{W}^{k}(i,:)}\hspace{-1em}\mathbf{W}^{k}(i,j) \mathbf{x}^{k-1}(j)
\end{split}
\label{equ:ip}
\end{equation}
which corresponds to the computation of the $i$th neuron's total input activation.
In the backpropagation phase, vertex $v_i$ represents column $(\mathbf{W}^{k})^T\!(:,i)$ and the task of computing multiplications in sparse SAXPY/DAXPY operations $\mathbf{s}^k(j)\!=\!\mathbf{s}^k(j)\!+\!(\mathbf{W}^{k})^T\!(j,i)\boldsymbol{\delta}(i)^{k}$ for each nonzero row index $j$ in column $(\mathbf{W}^{k})^T\!(:,i)$.
Additionally, vertex $v_i$ also represents gradient update operations 
\begin{eqnarray}
\textbf{W}^{k}(i,:) \leftarrow \textbf{W}^{k}(i,:) - \eta \nabla\textbf{W}^{k}(i,:) \label{equ:bp2}
\end{eqnarray}
\noindent associated with the links connected to the $i$th neuron in layer $k$.
Therefore, each vertex is associated with a computational weight equal to the number of nonzeros in row \mbox{$\mathbf{W}^{k}(i,:)$}~(i.e., number of links connecting to the $i$th neuron).
Fixed vertices in set $F^{k}$ do not represent any computation and are introduced to connect nets to prespecified parts for correctly encoding input-output dependencies between consecutive layers in multi-phase partitioning framework.

A $P$-way partitioning $\Pi_{P}(\phi^{k})\!=\!\{V^{k}_1,V^{k}_2,\ldots,V^{k}_{P}\}$ on hypergraph $H(\phi^{k})$ denotes that all tasks corresponding to vertices in part $V^{k}_m\!\in\!\Pi_{P}(\phi^{k})$ are assigned to processor $P_{m}$.
For instance, if a vertex $v_i$ is assigned to part $V^{k}_m$, then processor $P_m$ stores row $\mathbf{W}^{k}(i,:)$ and performs all computation associated with this row.
Partitioning $\Pi_{P}$ induces a partial reordering so that the matrix rows belonging to the same part can be reordered consecutively~(in any order) to form a row block $\mathbf{W}^{k}_{m}$  which is assigned to processor $P_m$.

Net set $N^{k}$ simultaneously encodes the total communication volume of processors during inference and backpropagation phases.
In the inference phase, each net $n_j\!\in\!N^{k}$ represents the set of tasks~(vertices) that need nonzero entry $\mathbf{x}^{k-1}(j)$, whereas in the backpropagation phase, each net $n_j$ represents the set of tasks that contribute to the computation of nonzero entry $\mathbf{s}^{k}(j)$.
Hence, net $n_j$ connects each vertex $v_i\!\in\!V^{k}$ for which the corresponding row $\mathbf{W}^{k}(i,:)$ has a nonzero entry in the $j$th column.

In order to satisfy input-output dependencies between successive layers, each net $n_j\!\in\!N^{k}$ connects only one fixed vertex $v^f_j\!\in\!F^{k}$ and fixed vertex $v^f_j$ only connects $n_j$.
Fixed vertex $v^f_j$ represents nonzero $\mathbf{x}^{k-1}(j)$ and it is fixed to the same part/processor to which row $\mathbf{W}^{k-1}(j,:)$ is assigned in the preceding phase $\phi^{k-1}$, since $\mathbf{x}^{k-1}(j)$ is locally computed by that processor in layer $k\!-1\!$~(i.e., $v_j\!\in\!V^{k-1}_m\!\Rightarrow\!v^f_j\!\in\!V^{k}_m$).
In other words, fixed vertex $v^f_j\!\in\!F^{k}$ ensures that after partitioning in phase $\phi^{k}$, net $n_j\!\in\!N^{k}$ connects the part/processor which is given the responsibility of computing nonzero $\mathbf{x}^{k-1}(j)$.
Formally, pins of net $n_j$ is defined as
\begin{equation}
\mathrm{pins}(n_j)= \{v_i \in V^k \mid \exists j \in \mathrm{rows}(\mathbf{W}^{k}(i,:))\} \cup \{v^f_j\}.
\end{equation}

In the inference phase, a cut net $n_j\!\in\!N^{k}$ whose fixed vertex $v^f_j$ is assigned to a part $V^{k}_m\!\in\!\Lambda(n_j)$ implies that nonzero $\mathbf{x}^{k-1}(j)$ is computed by processor $P_m$ and will be sent from $P_m$ to all processors in $\Lambda(n_j)\!\setminus\!V^{k}_m$.
Therefore, cut net $n_j$ incurs the communication volume of $\left(|\Lambda(n_j)|\!-\!1\right)$ words in the $k$th layer of SpFF.
In the backpropagation phase, each processor \mbox{$P_n\!\in\!\Lambda(n_j)\!\setminus\!V^{k}_m$}, computes its contribution to nonzero $\mathbf{s}^{k}(j)$ and sends to processor $P_m$.
Therefore,  cut net $n_j$ incurs the communication volume of $\left(|\Lambda(n_j)|\!-\!1\right)$ words in the backpropagation phase as well.
As seen here, if a processor $P_m$ sends a nonzero $\mathbf{x}^{k-1}(j)$ to a processor $P_n$ in the inference phase, processor $P_m$ receives the corresponding gradient contribution $\mathbf{s}^{k}(j)$ from processor $P_n$.
Therefore, the total communication volume between processors during SpFF and SpBP in layer $k$ can be given as
\begin{equation}
\mathrm{Vol}(k) = \sum\limits_{n_j \in N^{k}} 2 \times \left( |\Lambda(n_j)|-1 \right). \nonumber
\end{equation}
Therefore, if each net $n_j\!\in\!N^{k}$ is associated with \mbox{$cost(n_j)\!=\!2$}, the partitioning objective of minimizing the cutsize in phase $\phi^{k}$ encodes the minimization of the total communication volume during performance of SpFF and SpBP in layer $k$.
Note that each net is associated with equal \mbox{$cost(n_j)\!=\!2$} which encodes the number of nonzeros transferred during inference and backpropagation phases. Any uniform cost association is valid for partitioning.

\begin{figure}[t]
\centering
\includegraphics[width=0.5\textwidth]{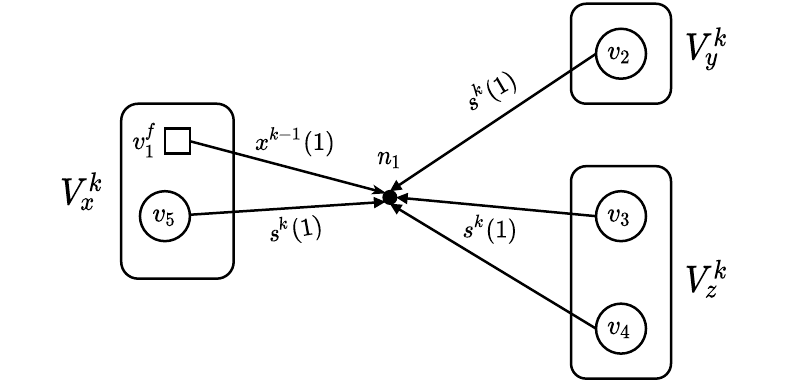}
\caption{Cut net $n_1$ encoding communication operations between neurons represented by $v^f_1$ in layer $k\!-\!1$ and neurons $v_2,v_3,v_4,v_5$ in layer $k$.}
\label{fig-net}
\end{figure}

Figure~\ref{fig-net} displays an illustrative example where cut net $n_1$ with $\Lambda(n_1)\!=\!\{V^k_x, V^k_y, V^k_z\}$ is given.
In the figure, fixed vertex $v^f_1$ is preassigned to part $V^k_x$ by partitioning $\Pi(\phi^{k-1})$ so that the task of computing $\mathbf{x}^{k-1}(1)$ is given to processor $P_x$~(i.e., $v_1\!\in\!V^{k-1}_x$).
Hence, in the $k$th step of inference phase,  processor $P_x$ sends $\mathbf{x}^{k-1}(1)$ to processors $P_y$ and $P_z$, since output of the neuron $v^f_1$ is connected to neurons $v_2$, $v_3$ and $v_4$.
Here neuron $v_5$ does not incur communication since it is assigned to the same processor $P_x$ by partitioning $\Pi(\phi^{k})$.
Even though the output of neuron $v^f_1$~(i.e., neuron $v_1$ in layer~$k\!-\!1$) is connected to two neurons $v_3$ and $v_4$ in processor $P_y$, nonzero $\mathbf{x}^{k-1}(1)$ is sent only once to this processor.
So net $n_1$ encodes a communication volume of $|\Lambda(n_j)|-1\!=\!2$ words during SpFF in layer $k$.
Similarly, in the backpropagation phase, processors $P_y$ and $P_z$ send partial gradient contributions for $\mathbf{s}^{k}(1)$ to processor $P_x$.
Partial gradient contribution of vertex $v_5$ is locally summed by $P_x$ and does not contribute to the total communication volume.
Note that $P_z$ sums partial gradients contributions for each of its vertices $v_3$ and $v_4$ before sending a single value to $P_x$.
Hence, as in the inference phase, net $n_1$ encodes the same communication volume of $|\Lambda(n_j)|-1\!=\!2$ words during SpBP in layer $k$.

\begin{figure*}[t]
\centering
\includegraphics[width=0.75\textwidth]{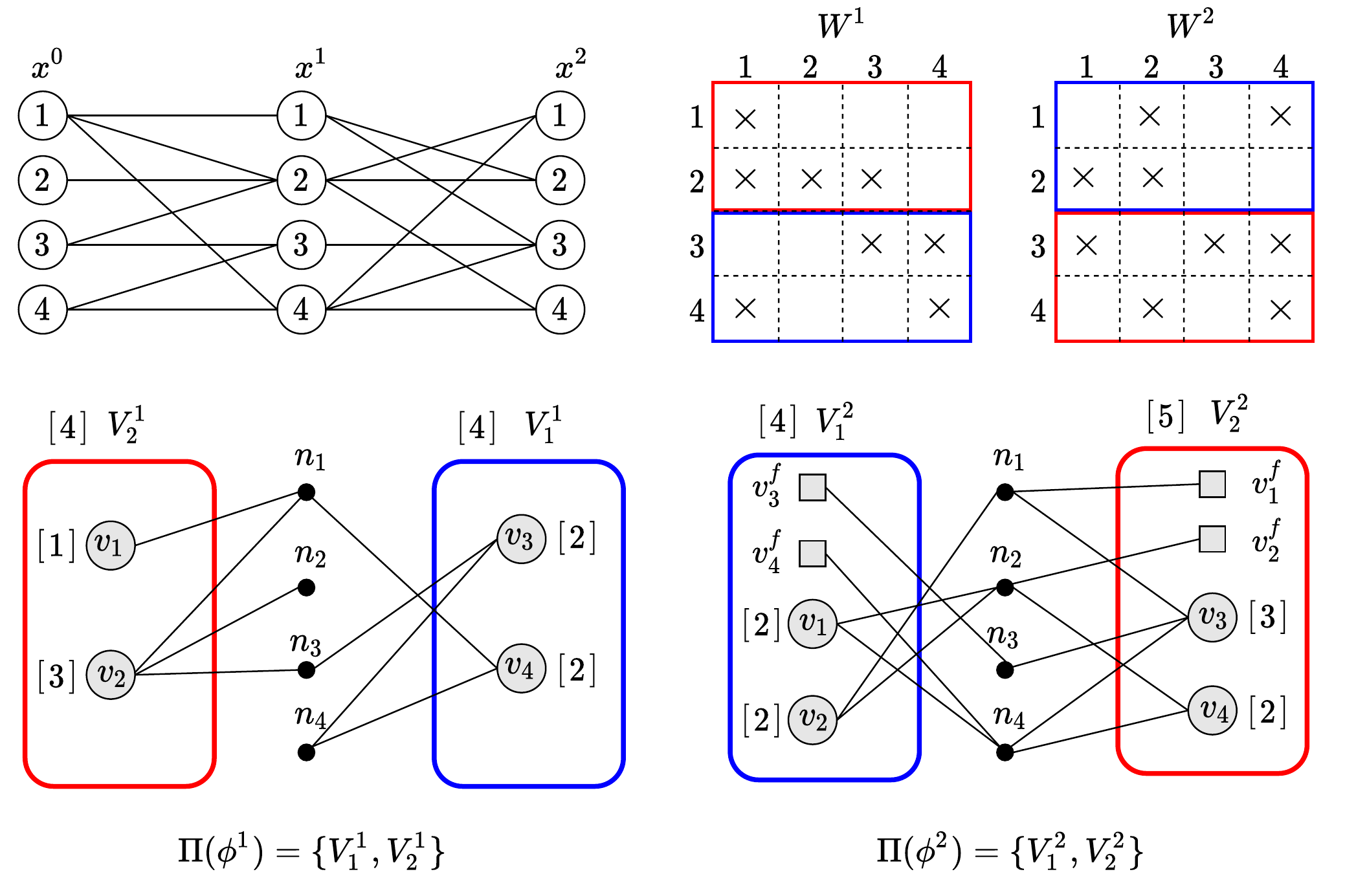}
\caption{Top: A 3-layer sparse DNN and the corresponding weight matrices $\mathrm{W}^{1}$ and $\mathrm{W}^{2}$. Bottom: Hypergraphs $H(\phi^{1})$ and $H(\phi^{2})$ built for weight matrices. Partitions $\Pi(\phi^{1})$ and $\Pi(\phi^{2})$ induce a 2-way partitioning on weight matrices. $[\cdot]$ next to a vertex or a part shows the weight associated with that vertex and part, respectively.}
\label{fig1}
\end{figure*}

Figure~\ref{fig1} displays an illustrative example of the proposed hypergraph partitioning model.
The sparse DNN in the top left in the figure consists of three layers each of which contains four neurons~(i.e., $\textbf{x}^{0}$ corresponds to the input layer).
Weight matrices $\textbf{W}^{1}$ and $\textbf{W}^{2}$ are displayed in the top right of the figure where connections between neurons are denoted by nonzero entries.
For instance, neuron~$2$ in the first layer is represented by row $\textbf{W}^{1}(2,:)$ where the columns $1,2$ and $3$ have nonzero entries, since neuron~$2$ connects neurons $1,2$ and $3$ in the input layer.
The two subfigures of the lower part display hypergraphs  $H(\phi^{1})$ and $H(\phi^{2})$ which contain four vertices and four nets corresponding to rows and columns of matrices $\textbf{W}^{1}$ and $\textbf{W}^{2}$, respectively.
Additionally, $H(\phi^{2})$ contains four fixed vertices which correspond to rows of $\textbf{x}^{1}$.
Fixed vertices $v^f_3$ and $v^f_4$ are preassigned to part $V^{2}_1$ whereas $v^f_1$ and $v^f_2$ are preassigned to part $V^{2}_2$, since $v_3$ and $v_4$ are assigned to $V^{1}_1$ whereas $v_1$ and $v_2$ are assigned to $V^{1}_1$ by $\Pi(\phi^{1})$ in the previous layer.
That is, nonzeros $\textbf{x}^{1}(3)$ and $\textbf{x}^{1}(4)$ are computed by the processor $P_1$ whereas the rows $\textbf{x}^{1}(1)$ and $\textbf{x}^{1}(2)$ are computed by the processor $P_2$.
Therefore, in layer $2$, nonzeros $\textbf{x}^{1}(3)$ and $\textbf{x}^{1}(4)$ will be sent from $P_1$ to $P_2$, and the rows $\textbf{x}^{1}(1)$ and $\textbf{x}^{1}(2)$ will be sent from $P_2$ to $P_1$.
In the figure, rows of the input vector $\textbf{x}^{0}$ can be assigned to processors with respect to net connectivities.
For instance, row $\textbf{x}^{0}(1)$ can be stored by one of the processors $P_1$ and $P_2$, since net $n_1$ connects both parts $V^{1}_1$ and $V^{1}_2$.
On the other hand, net $n_2$ only connects $V^{1}_2$ and hence, $\textbf{x}^{0}(2)$ is stored locally by $P_2$ and it is not communicated.

The running time complexity of the partitioning phase depends on the sizes of hypergraphs built in each phase and the partitioning algorithm/tool used. 
The sizes of hypergraphs are all linear in the number of rows, columns and nonzero entries of weight matrices in each layer.
Hence, the complexity of generating hypergraph $H(\phi^{k})$ for layer $k$ can be given as $\theta(N\!+\!nnz(\textbf{W}^{k}))$, where $N$ and $\mathrm{nnz}(\cdot)$ respectively denote the number of neurons per layer~(i.e., the number of rows and columns of $\textbf{W}^{k}$) and number of nonzero entries~(i.e., connections) in a matrix.

\subsection{Discussion}

One challenge inherent in the parallel SGD algorithm is that processors perform communication between each consecutive layer, introducing a synchronization barrier.
To alleviate synchronization overheads and improve the parallelization efficiency, input vectors can be processed in batches at each iteration~(i.e., minibatch SGD can be performed instead of SGD).
By simply modifying SpFF, batch processing can be enabled in such a way that instead of forwarding a single vector $\textbf{x}^{k}$ between each consecutive layer, multiple vectors can be simultaneously processed in batches.
That is, sparse matrix-matrix multiplications~(SpMM) of the form $\textbf{W}^{k}\textbf{X}^{k-1}$ can be performed in each layer where $\textbf{X}^{k-1}$ is formed by placing multiple $\textbf{x}^{k-1}$ vectors as columns in $\textbf{X}^{k-1}$.
Hence, the main iteration of the inference step becomes $\textbf{X}^{k}\!=\!f(\textbf{W}^{k} \textbf{X}^{k-1})$. 
The gradient vector $\boldsymbol{\delta}^{L}$ in the final layer is computed as the averages of gradients obtained over the vectors in the current batch.
The SpBP algorithm is executed in the same way, since a single gradient vector is backpropagated to update weight parameters.
Additionally, the proposed hypergraph partitioning is still applicable without any modifications, since the proposed model depends only on the DNN network structure.

The proposed hypergraph partitioning model can also be utilized for hybrid systems that provide both shared- and distributed-memory parallelism such as GPU or multiprocessor clusters.
Implementations that utilize ``MPI+CUDA'' or ``MPI+Openmp'' can benefit from the proposed hypergraph partitioning approach to reduce communication costs between compute nodes that are connected by slower network connections.
In this respect, our local SpMV computations can be replaced by more efficient libraries that utilize thread-level parallelism in multiprocessor and GPU architectures~\cite{bisson2019gpu,davis2019write}.
Additionally, the proposed hypergraph models can also be utilized for heterogeneous computation systems by enforcing different target part weights to distribute different sized computational loads to processors.

The proposed hypergraph partitioning model and the SpMV-based SGD can also be utilized for convolution/pooling layers, which are widely utilized in popular convolutional neural network~(CNN) architectures.
These layers can be implemented as matrix-vector multiplications through constructing Toeplitz matrices~\cite{gray2006toeplitz}, that capture convolution operation, and converting input data to vectors.
Application of sparsification/pruning to CNNs induces sparsification on the corresponding Toeplitz matrices, making the proposed hypergraph model applicable to such cases.

\section{Experiments}\label{sec-Performance}

\subsection{Experimental Setup}

We evaluate the performance of the proposed parallel SGD algorithm and hypergraph partitioning model on a benchmark provided by Sparse Deep Neural Network Graph Challenge\footnote{\url{https://graphchallenge.mit.edu/data-sets}}~\cite{kepner2019sparse}.
The benchmark uses synthetically generated sparse DNN models and MNIST database of handwritten digits~\cite{lecun1998mnist}.
These sparse networks are shown to be effective in terms of their training performance~\cite{kepner2019radix,prabhu2018deep}.
We refer to the parallel training algorithm as H-SGD if the proposed hypergraph partitioning model is used to partition the neural networks.
Otherwise, we refer to the algorithm as SGD to denote that random partitioning is utilized where neurons are assigned to processors uniformly at random in each layer.
Random partitioning evenly splits weight matrices by assigning rows to processors uniformly at random and provides competitive computation/communication balance.

Sparse DNNs are generated by RadiX-Net synthetic sparse DNN generator~\cite{kepner2019radix} which takes two parameters: the number of layers and the number of neurons per layer.
We used four different sized sparse DNNs consisting of $120$ layers where numbers of neurons per layer are selected as $N\!=\!1024,4096,16384,$ and $65536$, respectively.
The MNIST database consists of \mbox{60,000} images of size $28\!\times\!28$ pixels and these images are scaled to $32\!\times\!32$, $64\!\times\!64$, $128\!\times\!128$ and $256\!\times\!256$.
The scaled images are thresholded and flattened into 0-1 column vectors to be conformable with the input layers of sparse DNNs.

Running time experiments are performed on a high-performance computing system in which compute nodes are Lenovo NeXtScale nx360 M5 servers with 2$\times$Intel Xeon E5-2630 v3 2.4 GHz (Haswell) 8~core processors~(16~cores per node, 203 nodes, 3488 cores, 64GB DDR4 memory per node/4GB per core).
The system provides at most 32 compute nodes~(512 cores) to run our parallel codes. Compute nodes are connected via QLogic TrueScale InfiniBand.
To test the effectiveness of the proposed hypergraph partitioning model as well as the scalability of the parallel SpMV-based training algorithm, we performed strong scaling experiments for H-SGD and SGD on numbers of processors $P\!=\!32,64,128,256$ and $512$.
Our SGD algorithm currently supports single-thread execution where we assign a single core to each MPI process and run a single thread per MPI rank.
In our HPC system, the total memory of a compute node is not sufficient to store the whole DNN model for $N\!=\!65536$~(Data-parallel approaches fail due to memory constraints).
Therefore, our strong scaling experiments start from 32~cores~(i.e., 2~nodes).

We implemented the parallel sparse SGD algorithm in C++ and implemented the inter-process communication operations via Message Passing Interface~(MPI).
We used sigmoid function as linear activation function $f(\cdot)$ and mean squared error as loss function $\mathbf{J}$.
Initial connection weights of sparse DNNs are chosen uniformly at random from the interval $[-1,1]$ and the learning rate is set to $\eta\!=\!0.01$.
The proposed hypergraph model is partitioned by using Patoh~\cite{catalyurek1999hypergraph} where the maximum allowed imbalance ratio is set to $\epsilon\!=\!0.01$ in each layer.

Algorithms that only perform inference computations on sparse DNNs~\cite{mofrad2020studying,hidayetouglu2020scale,bisson2019gpu,davis2019write,pawlowski2020combinatorial} are not applicable in our general experimental setting. The best performing sparse DNN inference algorithms are generally designed for GPU-based systems and adopt data-parallelism. In these solutions, the backpropagation phase and weight update operations are not implemented. Data-parallel SGD solutions independently process input vectors in parallel and can not parallelize the computations associated with a single input vector, which limits the scalability by the batch size in training. 
Our solution achieves model-wise parallelism and can process a single input vector in parallel.
Due to this fundamental difference of objectives and functionalities, we omit comparison against data-parallel solutions.
To the best of our knowledge, our SGD solution is the first parallel SpMV-based training algorithm that achieves model-wise parallelism to train sparse DNNs on high-performance computing systems.

\begin{table}[t]
\caption{Performance comparison of SGD and H-SGD}
\label{tables:table1}
\centering
\setlength\tabcolsep{1.7pt}
\begin{tabular}{ll*5r||*5r}
\hline
\multicolumn{1}{c}{} & \multicolumn{1}{c}{} & \multicolumn{5}{c}{1024} & \multicolumn{5}{c}{4096} \\
\cmidrule(r){3-7}\cmidrule(r){8-12}
\multicolumn{1}{c}{} & \multicolumn{1}{c}{} & \multicolumn{2}{c}{Volume} & \multicolumn{2}{c}{Messages} & \multicolumn{1}{c}{} & \multicolumn{2}{c}{Volume} & \multicolumn{2}{c}{Messages} & \multicolumn{1}{c}{} \\
\cmidrule(r){3-4}\cmidrule(r){5-6}\cmidrule(r){8-9}\cmidrule(r){10-11}
\multicolumn{1}{c}{$P$} & \multicolumn{1}{c}{} & \multicolumn{1}{c}{Avg} & \multicolumn{1}{c}{Max} & \multicolumn{1}{c}{Avg} & \multicolumn{1}{c}{Max} & \multicolumn{1}{c}{imb} & \multicolumn{1}{c}{Avg} & \multicolumn{1}{c}{Max} & \multicolumn{1}{c}{Avg} & \multicolumn{1}{c}{Max} & \multicolumn{1}{c}{imb} \\
\hline
\multirow{3}{*}{$32$} 
&\multirow{2}{*}{H} 
&0.34 &0.34 &0.96 &0.96 &  &0.22 &0.22 &0.94 &0.94 &   \\
&&50 &52 &7 &7 &1.01 &130 &134 &7 &7 &1.01  \\
&\multirow{1}{*}{R} 
&149 &154 &7 &7 &1.05 &594 &603 &7 &7 &1.04  \\
\hline
\multirow{3}{*}{$64$} 
&\multirow{2}{*}{H} 
&0.31 &0.32 &0.82 &0.83 &  &0.23 &0.23 &0.93 &0.93 &    \\
&&29 &31 &12 &12 &1.01 &84 &87 &14 &14 &1.01 \\
&\multirow{1}{*}{R} 
&94 &98 &15 &15 &1.05 &375 &388 &15 &15 &1.05  \\
\hline
\multirow{3}{*}{$128$} 
&\multirow{2}{*}{H} 
&0.29 &0.28 &0.54 &0.55 &  &0.23 &0.23 &0.75 &0.76 &    \\
&&15 &16 &13 &14 &1.01 &48 &50 &23 &23 &1.01 \\
&\multirow{1}{*}{R} 
&53 &57 &24 &25 &1.08 &212 &222 &30 &30 &1.08  \\
\hline
\multirow{3}{*}{$256$} 
&\multirow{2}{*}{H} 
&0.39 &0.36 &0.49 &0.48 &  &0.20 &0.20 &0.42 &0.43 &    \\
&&9 &10 &11 &12 &1.03 &23 &24 &21 &22 &1.03 \\
&\multirow{1}{*}{R} 
&23 &26 &22 &24 &1.17 &113 &119 &51 &52 &1.17 \\
\hline
\multirow{3}{*}{$512$} 
&\multirow{2}{*}{H} 
&0.62 &0.53 &0.64 &0.58 &  &0.25 &0.25 &0.33 &0.34 &    \\
&&6 &6 &9 &9 &1.05 &12 &13 &15 &17 &1.05 \\
&\multirow{1}{*}{R} 
&10 &12 &14 &16 &1.24 &47 &52 &46 &49 &1.24 \\

\hline								
\multicolumn{1}{c}{} & \multicolumn{1}{c}{} & \multicolumn{5}{c}{16384} & \multicolumn{5}{c}{65536} \\
\hline					
\multirow{3}{*}{$32$} 
&\multirow{2}{*}{H} 
&0.17 &0.17 &0.92 &0.93 &  &0.15 &0.15 &0.91 &0.91 &  \\
&&407 &412 &7 &7 &1.01 &1,439 &1,454 &7 &7 &1.01  \\
&\multirow{1}{*}{R} 
&2,365 &2,377 &7 &7 &1.05 &9,419 &9,454 &7 &7 &1.05\\
\hline
\multirow{3}{*}{$64$} 
&\multirow{2}{*}{H} 
&0.16 &0.16 &0.92 &0.93 &  &0.13 &0.13 &0.91 &0.91 &   \\
&&240 &245 &14 &14 &1.01 &786 &796 &14 &14 &1.01 \\
&\multirow{1}{*}{R} 
&1,491 &1,512 &15 &15 &1.05 &5,938 &5,973 &15 &15 &1.05\\
\hline
\multirow{3}{*}{$128$} 
&\multirow{2}{*}{H} 
&0.15 &0.16 &0.90 &0.90 &  &0.12 &0.13 &0.91 &0.91 &   \\
&&130 &134 &27 &27 &1.01 &417 &424 &27 &28 &1.02  \\
&\multirow{1}{*}{R} 
&842 &859 &30 &30 &1.08 &3,355 &3,386 &30 &30 &1.08 \\
\hline
\multirow{3}{*}{$256$} 
&\multirow{2}{*}{H} 
&0.15 &0.15 &0.64 &0.65 &  &0.12 &0.12 &0.87 &0.88 & \\
&&69 &71 &39 &40 &1.03 &216 &221 &53 &53 &1.03  \\
&\multirow{1}{*}{R} 
&448 &462 &61 &61 &1.17 &1,786 &1,820 &61 &61 &1.17 \\
\hline
\multirow{3}{*}{$512$} 
&\multirow{2}{*}{H} 
&0.14 &0.14 &0.32 &0.33 &  &0.12 &0.12 &0.57 &0.58 & \\
&&32 &34 &33 &34 &1.05 &109 &112 &69 &70 &1.05  \\
&\multirow{1}{*}{R} 
&231 &243 &103 &105 &1.24 &922 &944 &122 &122 &1.24  \\
\hline
\end{tabular}
\end{table}

\subsection{Performance Results}

Table~\ref{tables:table1} compares the performance of SGD and H-SGD in terms of the communication volume and message counts metrics which relate to bandwidth and latency overheads of parallelization.
The table displays both the average and maximum volume/number of messages sent by a processor for comparison of the average and maximum values.
For each $P$, the first row displays the ratios of the respective values attained by H-SGD to those by SGD, whereas the second and third rows display actual values.
In the table, the last column shows the computational imbalance where the computational load is computed as the number of floating-point operations.

As seen in Table~\ref{tables:table1}, on all processor counts, H-SGD incurs 38--71$\%$, 75--80$\%$, 83--86$\%$ and 85--88$\%$ less average/total communication volume for sparse DNNs with $N\!=\!1024,4096,16384$ and $65536$, respectively.
Similarly, H-SGD incurs 47--72$\%$, 75--80$\%$, 83--86$\%$ and 85--88$\%$ less maximum send volume.
The decrease in the bandwidth-related costs increases as the size of DNNs increases.
The average and maximum communication volumes of processors are close to each other which denotes that communication balance is also achieved via the hypergraph partitioning.

\begin{figure}[t]
\centering
\begin{tikzpicture}[scale=0.7]

\begin{groupplot}[group style = {group size = 2  by 2, horizontal sep=1.1cm, vertical sep=2cm}, ticklabel style = {font=\large},width=6cm,height=6.5cm,xmode=log,ymode=log,xtick={ 32,64,128,256,512},xticklabels={ 32,64,128,256,512},grid=major]
    
    \nextgroupplot[very thick,title={\large $N\!=\!1024$},   xlabel= Processors, ylabel=Time~(msec), legend pos=north west,label style={font=\large},ytick={ 4,8,16,32,64,128},yticklabels={ 4,8,16,32,64,128}, ymin=4, legend columns=-1,legend style={at={(1.1,1.32)},anchor=north,font=\large}]
    \addplot[dashed] plot coordinates {(32,95.215245)(64,47.607622)(128,23.803811)(256,11.901906)(512,5.950953)};    
	\addplot[line width=1.3pt, mark=x,color=red,mark options={scale=2.5}] plot coordinates {(32,95.215245)(64,68.699137)(128,55.087236)(256,42.790836)(512,41.094719)};
	\addplot[line width=1.3pt, mark=x,color=blue,mark options={scale=2.5}] plot coordinates {(32,156.343535)(64,136.880155)(128,125.873613)(256,102.474583)(512,98.452104)	};
	
	\legend{Ideal,H-SGD,SGD};
        
    \nextgroupplot[very thick, title={\large $N\!=\!4096$},ytick={ 16,32,64,128,256,512},yticklabels={ 16,32,64,128,256,512}, xlabel= Processors, label style={font=\large}]
    \addplot[dashed] plot coordinates {(32,272.432581)(64,136.21629)(128,68.108145)(256,34.054073)(512,17.027036)};
	\addplot[line width=1.3pt, mark=x,color=red,mark options={scale=2.5}] plot coordinates {(32,272.432581)(64,192.612694)(128,135.572514)(256,99.420783)(512,81.558275)};
	\addplot[line width=1.3pt, mark=x,color=blue,mark options={scale=2.5}] plot coordinates {(32,572.766002)(64,396.53419)(128,321.295894)(256,278.512784)(512,239.331368)};

	\nextgroupplot[very thick, title={\large $N\!=\!16384$},ytick={ 64,128,256,512,1024,2048},yticklabels={ 64,128,256,512,1024,2048}, ylabel=Time~(msec), xlabel= Processors, label style={font=\large}]
	\addplot[dashed] plot coordinates {(32,1143.135619)(64,571.5678095)(128,285.78390475)(256,142.891952375)(512,71.4459761875)};
	\addplot[line width=1.3pt, mark=x,color=red,mark options={scale=2.5}] plot coordinates {(32,1143.135619)(64,570.166995)(128,344.227166)(256,259.488084)(512,199.472899)};
	\addplot[line width=1.3pt, mark=x,color=blue,mark options={scale=2.5}] plot coordinates {(32,2584.964416)(64,1499.925865)(128,997.39456)(256,775.086196)(512,670.60317)};

	\nextgroupplot[very thick, title={\large $N\!=\!65536$}, xlabel= Processors, label style={font=\large}, ytick={ 512,1024,2048,4096,8192,16384},yticklabels={ 512,1024,2048,4096,8192,16384}]
	\addplot[dashed] plot coordinates {(32,6624.711624)(64,3312.355812)(128,1656.177906)(256,828.088953)(512,414.0444765)};
	\addplot[line width=1.3pt, mark=x,color=red,mark options={scale=2.5}] plot coordinates {(32,6624.711624)(64,2912.757801)(128,1345.212811)(256,745.57489)(512,571.243823)};
	\addplot[line width=1.3pt, mark=x,color=blue,mark options={scale=2.5}] plot coordinates {(32,23791.295637)(64,7384.361263)(128,4120.877429)(256,2519.632116)(512,2142.731074)};

  \end{groupplot}
\end{tikzpicture}
\caption{Strong scaling of SGD and H-SGD on different sized sparse DNNs consisting of $120$ layers, and the number of neuron per layer $N=1024,4096,16384$ and $65536$, respectively.}
\label{fig:strong-scaling}
\end{figure}
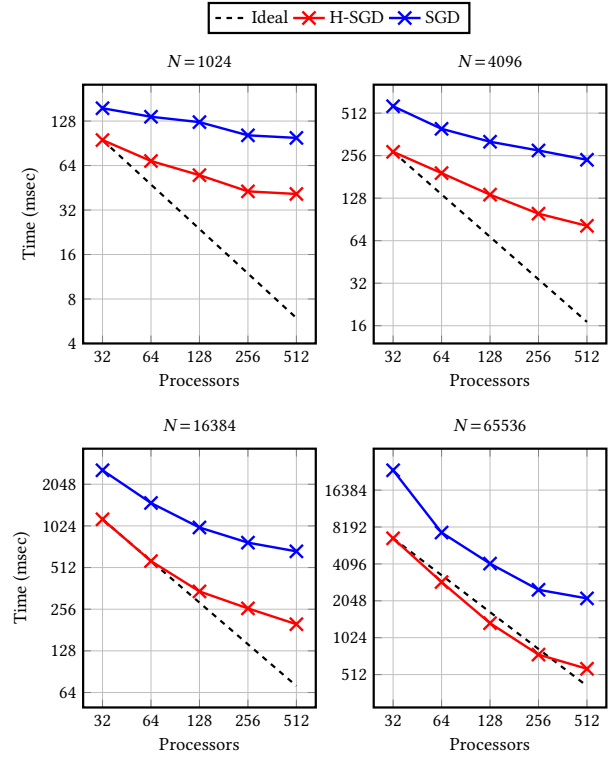

In terms of the message count metrics, H-SGD achieves 4--51$\%$, 6--67$\%$, 8--68$\%$ and 9--43$\%$ smaller average message counts and 4--52$\%$, 6--66$\%$, 7--67$\%$ and 9--42$\%$ smaller maximum message counts.
As the number of processors increases, the performance gap between the message count metrics of H-SGD and SGD increases in favor of H-SGD.
In terms of computational load balance, H-SGD provides consistently better performance than SGD.
These results demonstrate the effectiveness of the proposed hypergraph partitioning model since both the bandwidth- and latency-related costs are considerably minimized.
Moreover, as the number of layers in sparse DNNs increases, performance improvement of the hypergraph partitioning model is expected to be higher due to optimizations achieved in each layer.

Figure~\ref{fig:strong-scaling} shows strong scaling of SGD and H-SGD.
On each processor count, running times are measured as the average time required to process an input vector by H-SGD and SGD, where the averages are taken over  $10^3$ randomly selected input vectors.
For all processor counts, H-SGD considerably improves the parallelization efficiency and runs 2.01--2.37x, 1.97--2.96x, 2.10--3.39x and 2.88--3.37x faster than SGD on sparse DNNs with $N\!=\!1024,4096,16384$ and $65536$, respectively.
The best speedup is achieved on $N\!=\!65536$ and $P\!=\!512$ where H-SGD runs 3.39x faster than SGD.
H-SpBP achieves the ideal speedup up to $512$ processors on DNN with $N\!=\!65536$.

The synchronization barrier due to the communication operations between successive layers constitutes the main source of latency overheads of the parallel SGD algorithm.
As seen in Figure~\ref{fig:strong-scaling}, the efficiency of parallel SGD algorithm considerably improves with the increasing number of neurons per layer, since latency overheads are considerably amortized on larger networks.
Additionally, the performance improvement achieved on running time by hypergraph partitioning increases with the increasing sizes of DNNs as well as the increasing number of processors.

\begin{figure}[t]
\centering
\includegraphics[width=0.4\textwidth]{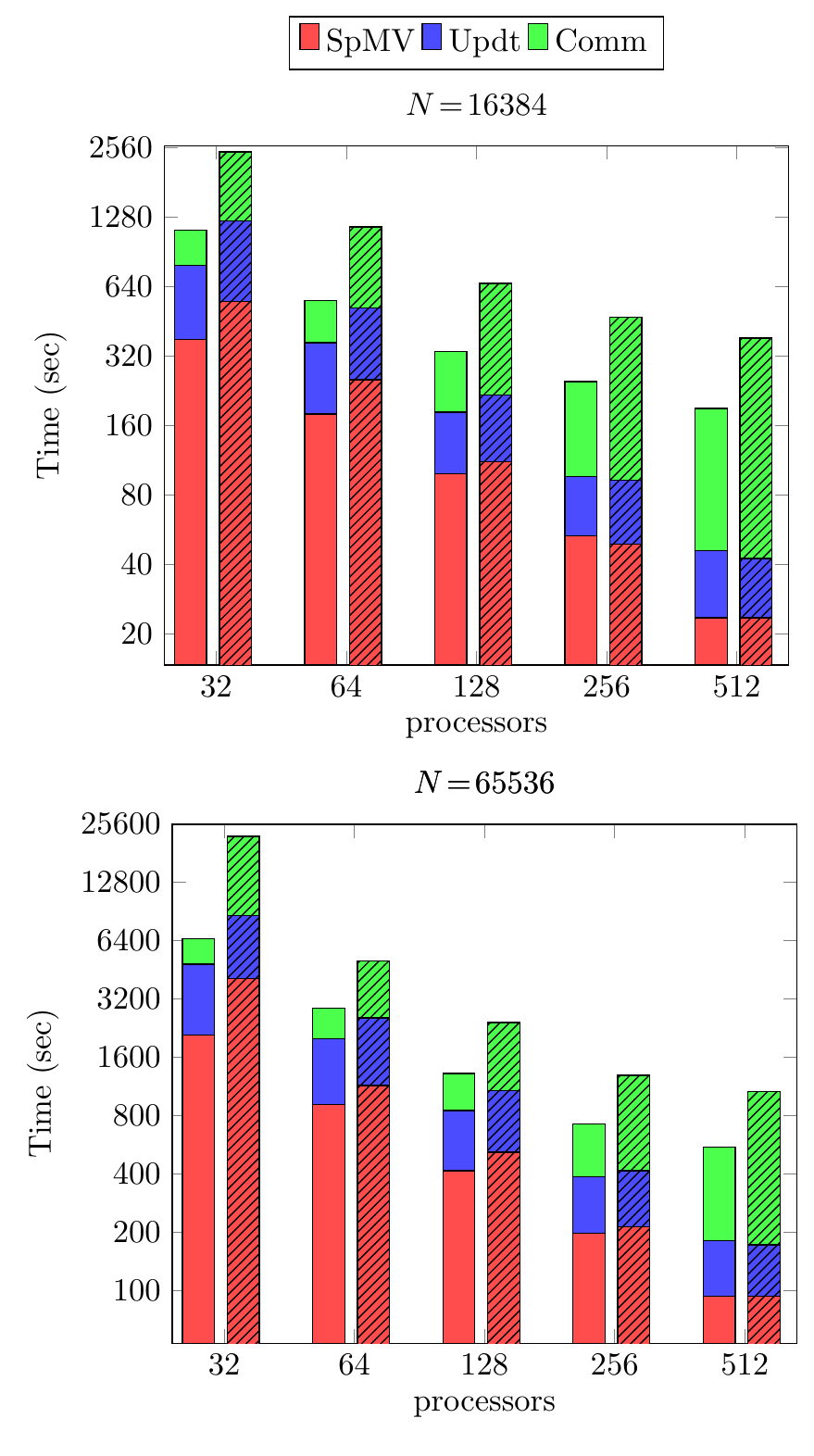}
\caption{Breakdown of running time of H-SGD~(solid) and SGD~(tiled). ``SpMV'' corresponds to time spent on local sparse matrix-vector multiplications.``Updt'' corresponds to the time spent on gradient update operations.``Comm'' corresponds time spend for communication operations.}
\label{fig:barchart}
\end{figure}

In Figure~\ref{fig:barchart}, to better analyze the effects of the hypergraph partitioning on the performance of SGD, we break down the total time spent on communication and computation.
As seen in the figure, the proportion of communication time to the overall running time increases with the increasing number of processors, whereas the proportion of time spent on local SpMV and gradient update computations decrease together with the total running time.
For example,  when $N\!=\!65536$, the proportion of communication time respectively increases from $26\%$ to $67\%$ and $40\%$ to $80\%$ for H-SGD and SGD as the number of processors increases from $p=32$ to $p=512$.
Hence, the improvements of hypergraph partitioning on the communication costs become more significant on the overall running time on larger processor counts.
As the number of processors increases, the ratio of improvement in communication time to the improvement in the overall execution time gradually increases from $48\%$ to $82\%$ and $50\%$ to $80\%$ for $N\!=\!16384$ and $N\!=\!65536$, respectively.
This can be attributed to the fact that on larger processor counts, communication costs become more dominant on the overall parallelization overheads and optimizations achieved by hypergraph partitioning on communication volume and message count metrics considerably improves.

We also observe that the hypergraph partitioning improves the performance of local SpMV and gradient update computations.
Specifically, H-SGD reduces the running time of local computations by $1.9$--$2.7$x on all processor counts as compared to SGD.
The performance improvement on the local computations arises because hypergraph partitioning consistently achieves better computational balance and temporal cache-locality than random partitioning.
The hypergraph partitioning assigns weight matrix rows, that are accessing similar input vector entries, to the same processor, which provides temporal cache locality in accessing input vector entries during local SpMV and gradient update computations.
We refer the reader to~\cite{akbudak2013hypergraph} for a detailed explanation of how temporal cache-locality is achieved.

\subsection{Inference-only Computations}

For inference-only computations, we enhanced SpFF by implementing local sparse matrix operations via SuiteSparse:GraphBLAS library~\cite{davis2019algorithm}. 
The enhanced SpFF implementation supports batch processing and multi-thread execution.
We also use the proposed hypergraph partitioning model and hence, we refer to SpFF as H-SpFF here. 
We compare H-SpFF against a data-parallel solution~(GB)~\cite{davis2019write}, that became one of the Graphchallange 2019 champions.
GB utilizes SuiteSparse:GraphBLAS library to achieve shared-memory parallelism and is able to run on a single compute node.
Similar to GB, H-SpFF processes all input vectors in a single batch.

\begin{table}[t]
\caption{Throughput results}
\label{tables:table2}
\centering
\begin{tabular}{rr*5r}
\multicolumn{2}{c}{} & \multicolumn{1}{c}{H-SpFF} & \multicolumn{2}{c}{GB} \\
\cmidrule(r){3-3}\cmidrule(r){4-5}\cmidrule(r){6-7}
\multicolumn{1}{c}{Neurons} & \multicolumn{1}{c}{Layers} & \multicolumn{1}{c}{Throughput} & \multicolumn{1}{c}{Throughput} & \multicolumn{1}{c}{Speedup}\\
\hline
\multirow{3}{*}{1024}  & 120  & 4.90E+10 & 7.11E+10 & 0.69  \\
                       & 480  & 5.41E+10 & 8.55E+10 & 0.63  \\
                       & 1920 & 5.57E+10 & 8.89E+10 & 0.63  \\
\hline
\multirow{3}{*}{4096}  & 120  & 3.87E+10 & 7.38E+10 & 0.52  \\
                       & 480  & 3.71E+10 & 8.58E+10 & 0.43  \\
                       & 1920 & 3.63E+10 & 8.70E+10 & 0.42  \\
\hline
\multirow{3}{*}{16384} & 120  & 8.20E+10 & 5.13E+10 & 1.60  \\
                       & 480  & 7.91E+10 & 5.60E+10 & 1.41  \\
                       & 1920 & 7.81E+10 & 5.61E+10 & 1.39  \\
\hline
\multirow{3}{*}{65536} & 120  & 9.01E+10 & 2.80E+10 & 3.21  \\
                       & 480  & 8.57E+10 & 2.85E+10 & 3.01  \\
                       & 1920 & 8.55E+10 & 2.85E+10 & 3.00  \\
\hline
\end{tabular}
\end{table}

Table~\ref{tables:table2} compares throughput values achieved by H-SpFF and GB for all sparse DNN configurations.
Throughput corresponds to the ratio of the number of input vectors times the number of connections in a DNN divided by the execution time~(i.e., number of edges processed per second).
The best throughput values of H-SpFF are measured on $512$ cores with $128$ MPI processes where we assign $4$~cores for each MPI process and run $4$~threads per MPI rank.
We run GB on a single node in our local HPC system where the last two columns in the table display throughput and the relative speedup values measured on our local system.
Standard nodes' memories were not enough for GB; hence we used fat nodes, which are in less number,  that contain the same CPU configuration with higher memory.

As seen in Table~\ref{tables:table2}, H-SpFF performs slightly worse than GB for small networks, whereas its performance considerably improves for larger networks, providing higher speedup values.
For network configurations with $N\!=\!16384,65536$ and $L\!=\!120$, H-SpFF achieves $1.6$x and $3.2$x speedups over GB, respectively.
This can be attributed to the fact that the latency overheads introduced by the synchronization barrier between successive layers reduce the parallelization efficiency. 
The latency overheads are considerably amortized as the number of neurons per layer increases and the number of layers decreases.
Therefore, H-SpFF is expected to perform better for network configurations with higher number of neurons and lower number of layers.

\subsection{Partitioning Times}

The preprocessing overhead of the partitioning is easily amortized, since the partitioning overhead is independent of the number of input vectors~(i.e., training data size) fed into sparse DNNs, whereas the communication costs and the performance improvement attained by the hypergraph partitioning model increases with the increasing number of input vectors.
Partitioning is performed once for each layer. 
Sets $\mathrm{Xsend}$ and $\mathrm{Xrecv}$ are computed in partitioning time and not modified hence do not affect the runtime.
Table~\ref{tables:table3} displays partitioning times for $L\!=\!120$ layer sparse DNNs we used in our experiments.
As seen in the table, as the number of parts and the number of neurons per layer increases, partitioning times increase.
Partitioning times are measured on a server with \textit{2$\times$Intel~Xeon~W-2245} 3.90GHz 8~core processors and 500GB DDR4 main memory.

\begin{table}[t]
\caption{Partitioning times~(secs)}
\label{tables:table3}
\centering
\begin{tabular}{r*4r}
$P$ &	1024 & 4096 & 16384 & 65536 \\
\hline
32 &	2.48 &	10.93 &	52.61 &	344.79 \\
64 &	3.41 &	12.57 &	63.09 &	355.03 \\
128 &	3.89 &	13.46 &	67.46 &	387.56 \\
256	&   4.97 &	16.77 &	71.59 &	408.48 \\
512	&   5.63 &	20.85 &	77.91 &	423.17 \\
\hline
\end{tabular}
\end{table}

\section{Conclusion}\label{sec-conclusion}

We first introduced a distributed-memory parallel sparse DNN inference/training algorithm for high-performance computing systems.
The solution is based on efficient parallelization of consecutive SpMV operations and achieves model-wise parallelism which significantly eliminates memory and bandwidth bottlenecks inherent in data-parallel approaches.
We then proposed a novel hypergraph partitioning-based solution
to address the latency overheads due to the communication operations between consecutive layers. The hypergraph partitioning model considerably improves communication overheads by reducing the total communication volume and the number of messages between processors while satisfying computational balance.
Extensive experiments suggest that the proposed model-wise parallel solution scales to large processor counts especially when the proposed hypergraph partitioning is utilized.
With the increasing number of neurons per layer and decreasing number of layers, latency overheads between consecutive layers are considerably amortized.
Therefore, in cases where the whole DNN model can not fit into main memory and the data-parallel approaches are not feasible, the model-wise parallel inference/training algorithm and hypergraph partitioning model offer a feasible alternative for distributed memory systems.

\section{Acknowledgments}

Computing resources used were provided
by The Scientific Computing Research Technology Platform\footnote{\url{https://warwick.ac.uk/research/rtp/sc/}} at University of Warwick.

\bibliographystyle{ACM-Reference-Format}
\bibliography{paper}

\end{document}